\documentclass[authoryear,preprint,11pt]{elsarticle}

\usepackage{amssymb}
\usepackage{natbib}
\usepackage{dirtytalk}  %% for quotations
\usepackage{hyperref}
\usepackage{makecell}  %% used for formatting cells in the table
\usepackage{float}  %% better placement of floating objects
\restylefloat{table}
\usepackage{caption}  %% allows side by side figures and captions
\usepackage{subcaption}  %% ditto
\usepackage[table]{xcolor}  %% can colour text in tables
\usepackage{soul}  %% provides text highlighting

\usepackage{algorithm} %% describes algorithms
\usepackage{algpseudocode}  %% algorithm pseudo code

\journal{arXiv}

\begin{document}

\begin{frontmatter}



\title{Causal discovery using dynamically requested knowledge}


\author[qmul]{Neville K. Kitson\corref{cor1}}
\ead{n.k.kitson@qmul.ac.uk}
\cortext[cor1]{Corresponding author}
\author[qmul]{Anthony C. Constantinou}
\ead{a.constantinou@qmul.ac.uk}

\affiliation[qmul]{organization={Bayesian Artificial Intelligence Research Lab, Machine Intelligence and Decision Systems (MInDS) Group, Queen Mary University of London (QMUL)}, 
                   city={London}, postcode={E1 4NS}, country={United Kingdom}}

\begin{abstract}
Causal Bayesian Networks (CBNs) are an important tool for reasoning under uncertainty in complex real-world systems. Determining the graphical structure of a CBN remains a key challenge and is undertaken either by eliciting it from humans, using machine learning to learn it from data, or using a combination of these two approaches. In the latter case, human knowledge is generally provided to the algorithm before it starts, but here we investigate a novel approach where the \textit{structure learning algorithm itself} dynamically identifies and requests knowledge for relationships that the algorithm identifies as \say{uncertain} during structure learning. We integrate this approach into the Tabu structure learning algorithm and show that it offers considerable gains in structural accuracy, which are generally larger than those offered by existing approaches for integrating knowledge. We suggest that a variant which requests only arc orientation information may be particularly useful where the practitioner has little preexisting knowledge of the causal relationships. As well as offering improved accuracy, the approach can use human expertise more effectively and contributes to making the structure learning process more transparent.
\end{abstract}


\end{frontmatter}


\section{Introduction}
\label{sect:introduction}

Causal Bayesian Networks (CBNs) provide a potentially powerful means of understanding and intervening in complex real-world systems as discussed in, for example, \cite{koller2009probabilistic} and \cite{darwiche2009modeling}. \cite{pearl2018book} describe the reasoning capabilities of models in terms of a \say{ladder of causation} with three levels of increasing reasoning power. Most Artificial Intelligence (AI) models today provide only Level 1 capabilities, that is, predictive capabilities. In contrast, CBNs provide Level 2 and 3 abilities to model the effect of interventions and to perform counterfactual reasoning, respectively. Moreover, CBNs represent the causal relationships as an intuitive graphical structure and thus provide transparency and explainability, something that is often absent in 'black box' models such as neural networks. These attractive features have meant that CBNs continue to be used to model complex real-world systems in a wide range of application domains such as healthcare \citep{sesen2013bayesian,shen2020challenges}, biology \citep{sachs2005causal,bernaola2020learning}, engineering \citep{cai2018application} and the environment \citep{graafland2020probabilistic,runge2019detecting}.

Notwithstanding the above, accurately determining the causal graphical structure which underlies a CBN remains a challenging problem. It may be specified by humans, a methodology referred to as \textit{knowledge elicitation}. This is generally performed using some formal framework such as Knowledge Engineering of Bayesian Networks (KEBN) described by \cite{korb2010bayesian}; although, as they and \cite{marcot2017common} point out, knowledge elicitation brings issues of human mistakes and biases. Alternatively, the graphical structure may be learnt from data using \textit{structure learning} algorithms which we discuss in the next section. A third approach is to combine machine learning and human knowledge which is often referred to as \textit{structure learning with knowledge} and is the topic of this paper.

Traditionally, human knowledge has been presented to the structure learning algorithm as \textit{predefined knowledge}\footnote{We deliberately use the term \textit{predefined} rather than \textit{prior} here as it specifies that any prior knowledge about the causal structure is supplied \textit{before} the algorithm starts. This contrasts with \textit{active learning} where that prior knowledge is provided \textit{during} the learning process.} before it begins the learning process, with no guidance from the algorithm as to what knowledge might be most beneficial. Less commonly, algorithms are used to specify which human knowledge might be most helpful in improving structure accuracy, an approach often referred to as \textit{active learning}. Our significant contribution is to suggest a novel approach whereby the \textit{structure learning algorithm itself} provides guidance as to where human knowledge might be most helpful in improving accuracy. We show that this approach is generally more effective than using predefined knowledge and so is an approach worthy of consideration when predefined knowledge is not available. We also examine the effect of inaccurate human knowledge in some detail.

The next section describes structure learning algorithms and how human knowledge may be integrated into the learning process. We cover both predefined knowledge and active learning approaches. Section~\ref{sect:implementation} describes how we modify the competitive and widely-used Tabu score-based algorithm to produce a new structure learning algorithm called Tabu-AL which implements active learning. Section~\ref{sect:evaluation} describes how the approach is evaluated with the results presented in Section~\ref{sect:results}, and concluding remarks are made in Section~\ref{sect:conclusions}.

\section{Background}
\label{sect:background}

\subsection{Bayesian Networks}

Bayesian Networks (BNs) were introduced by \cite{pearl1985bayesian} and are Probabilistic Graphical Models (PGMs) which use a Directed Acyclic Graph (DAG) to represent the probabilistic dependency relationships between variables. Each node in the DAG represents a variable under study and each arc represents a direct dependence relationship between the two variables it connects. BNs obey the Local Markov property which states that a child is conditionally independent of all other nodes given its parent nodes. This property implies that the standard chain rule for expressing a global probability distribution:

\begin{equation}
P(X_1, X_2, ..., X_n)=\prod_{i=1}^n P(X_i | X_1, X_2, ..., X_{i-1})
\end{equation}
can be expressed much more concisely as:

\begin{equation}
P(X_1, X_2, ..., X_n)=\prod_{i=1}^n P(X_i | \textbf{Pa}(X_i))
\end{equation}
where $X_1, X_2, ..., X_n$ are the $n$ variables in the BN, and $\textbf{Pa}(X_i)$ are the parent variables of $X_i$ in the BN. Thus a BN provides a compact way of expressing the global probability distribution of the variables.

The local Markov property gives rise to further useful properties of the DAG, in particular, the fact that a graphical property of the nodes in the DAG called \textit{d-separation} can be used to quickly establish whether two variables are conditionally independent of one another given any other set of variables. The BN also specifies the exact dependency relationship between each node and its parents (if any). This paper considers BNs that include only discrete variables and, in that case, each direct dependency relationship is defined by a Conditional Probability Table (CPT) which specifies the probability of the child variable taking each of its possible values depending upon each combination of values that its parents may take. Given the DAG and the CPTs, it is possible to compute the marginal or conditional probabilities of any subset of variables, a process usually referred to as \textit{causal inference} in this field.

It is possible to show that, in general, more than one BN, each with a different DAG, can give rise to the same global probability distribution \citep{verma1990equivalence}. This set of BNs is called a Markov Equivalent Class, and the DAGs in the class are described as being \textit{Markov equivalent} or, more simply, \textit{equivalent}. The set of equivalent DAGs can be represented by a Completed Partially Directed Acyclic Graph (CPDAG) that contains both undirected and directed edges. The directed edges in the CPDAG indicate edges that have the same orientation in all DAGs in the equivalence class, and the undirected edges indicate edges that have one orientation in some of the DAGs, and the other orientation in the others. 

Causal Bayesian Networks (CBNs) make a further assumption that the directed edges in the DAG represent causal relationships so that each parent is a direct cause of the child. This provides a causal understanding of the relationships between the variables and allows us to model the effects of interventions and undertake counterfactual reasoning as mentioned in Section~\ref{sect:introduction}. Note that it may not always be correct to assume that relationships are causal, but this assumption is required for causal modelling. The effects of interventions can be modelled in a CBN using the \textit{do-operator} \citep{pearl2012calculus}. This simulates interventions by \textit{graph surgery} whereby the direct causes of a variable are removed, and the value of this intervened variable is set to a specific desired value independent of its causes, and the effect on variables of interest is computed.

\subsection{Structure Learning Algorithms}
\label{sub:algos}

As noted in Section~\ref{sect:introduction}, the structure of a CBN may be elicited from people or one can use structure learning algorithms to learn about the causal structure from data. The two main machine learning approaches for discrete categorical data are constraint-based and score-based algorithms. Constraint-based approaches use statistical Conditional Independence (CI) tests to identify the dependence and independence relationships present in the data and use the properties of BNs to infer the graphical structure from these. Given the fact that a set of independence relationships is usually compatible with more than one DAG, constraint-based algorithms generally return a CPDAG. The PC algorithm \citep{spirtes1991algorithm} and a variant that is less sensitive to variable ordering, PC-Stable \citep{colombo2014order}, are two commonly used constraint-based algorithms.

Score-based algorithms assign a score to each DAG visited. Bayesian scores such as BDeu \citep{heckerman1995learning} indicate the most probable graph given the data and some prior beliefs about the structure. Information-theoretic scores such as BIC \citep{suzuki1999learning} balance the likelihood of the DAG generating the data against model complexity. Score-based algorithms search over graphs and return the highest-scoring graph they discover.

Exact score-based algorithms such as GOBNLIP \citep{cussens2011bayesian} and A-Star \citep{yuan2011learning} potentially guarantee to return the highest-scoring DAG out of all possible DAGs for the data set being learnt from. However, the extended runtimes that this involves means that, in practice, exact algorithms are restricted to considering DAGs with an upper limit on the number of parents each variable can have when used on problems with several tens of variables or more. Approximate score-based algorithms offer no guarantee that the returned graph is the highest-scoring one, but this does not imply that they will recover a less accurate causal structure, and makes the algorithms applicable to larger datasets containing hundreds or (depending on the algorithm) even thousands of variables. For example, the DAG hill-climbing (HC) algorithm \citep{bouckaert1992optimizing} is a simple approximate algorithm which starts from an empty DAG and greedily adds, removes or reverses an arc at each iteration until the score no longer increases. The Tabu algorithm \citep{bouckaert1995bayesian} is a variant of HC which allows iterations where the score decreases and is the algorithm we base the work in this paper. Some score-based algorithms such as GES \citep{chickering2002optimal} and its optimised variant FGES \citep{ramsey2017million} search through CPDAGs rather than DAGs and are notable in being greedy algorithms that guarantee to return the highest scoring CPDAG, but only if the sample size is large enough.

Most objective functions used, including BIC and BDeu, are \textit{score equivalent} meaning that they return the same value for all DAGs in an equivalence class, so that, just like constraint-based algorithms, they do not differentiate between equivalent DAGs. Nevertheless, most score-based algorithms do actually return a DAG rather than a CPDAG, but this DAG is generally best regarded as an example from the equivalence class of DAGs, and usually, its corresponding CPDAG is used when evaluating the result.

Other classes of structure learning algorithms have developed more recently. Hybrid algorithms use a combination of constraint and score-based approaches. For example, MMHC \citep{tsamardinos2006max} employs a commonly adopted strategy of using a constraint-based algorithm to define a more limited set of DAGs for a subsequent score-based approach to search within.

The algorithms discussed so far deal with discrete graphs during the learning process - that is, an edge either exists or does not exist at each step of the algorithm. Recently, continuous optimisation algorithms such as NOTEARS \citep{zheng2018dags} represent the graph in terms of a real-valued matrix whose elements ascribe a fractional value to the existence of each arc. This allows the structure learning problem to be tackled by off-the-shelf continuous optimisation approaches in a manner more akin to learning a neural network.

The equivalence class issue means that we are generally unable to learn a causal DAG from observational data alone. Recent approaches have attempted to address this issue by using interventional as well as observational data; for example, the COMBI \citep{triantafillou2015constraint} or mFGS-BS \citep{chobtham2023hybrid} algorithms. Another approach to identifying causal orientations is to make specific assumptions about the form of the noise elements of the dependency relations, which then allows a causal direction to be identified \citep{peters2014causal}.

Structure learning algorithms are usually evaluated in one of two ways depending upon whether a real-world data set or a synthetic data set is used. In the former case, the ground truth causal graph is generally not known, and so the learned graph is assessed by how well it explains the data set or by its predictive capabilities. Alternatively, evaluation is performed by using an assumed ground truth graph to generate a synthetic data set, learning a graph from that synthetic data set, and then comparing the learned graph to the data-generating graph. The arcs in the learned and data-generating graphs are compared, and a metric such as F1 or SHD \citep{tsamardinos2006max} is used to summarise the accuracy of the learned graph.

In general, most structure learning algorithms make a series of, very often, quite restrictive assumptions that mean that the learned graph may be a poor reflection of reality when graphs are learnt from real-world data. These assumptions can include that:
\begin{itemize}
    \item there are no independence relationships in the data which are not implied by the causal graph. The presence of such inconsistencies is referred to as \textit{unfaithfulness} and arises when the effects of two causal paths 'cancel out';
    \item the variables belong to well-known distributions such as the Gaussian for continuous variables and the multinomial for discrete variables;
    \item there is no missing data
    \item there are no latent confounders - that is, unobserved variables which are causes of two or more of the observed variables under study
    \item there are no measurement or discretisation errors
\end{itemize}

Progress has been made in addressing many of these assumptions. For example, the FCI \citep{spirtes2000causation} family of algorithms take account of latent confounders, the Structural EM approach \citep{friedman1997learning} can tackle missing data, and recent work by \cite{liu2022improving} corrects for measurement error. \cite{kitson2023survey} provide a comprehensive view of recent developments in structure learning algorithms.

Nonetheless, causal structure learning remains a challenging task even when the above assumptions \textit{are} satisfied. Limited sample sizes may mean that CI tests are unreliable, or the asymptotic assumptions on which many of the scores are based do not hold. Even if the sample size is sufficiently large, which in itself is not a well-defined limit, equivalence means that a fully identified causal graph cannot be learned from observational data.

The performance benchmark by \citep{scutari2019learns} indicates that learned graph accuracy is often modest and varies widely between different algorithms even under favourable assumptions such as no noise, and \cite{constantinou2021large} find that reasonable levels of data noise can halve structure learning accuracy. \cite{kyrimi2021bayesian} note that the uptake of BNs in production use in healthcare has been limited, including for reasons of limited accuracy. Thus, there is continued interest in using human knowledge to aid structure learning algorithms which we discuss next.

\subsection{Algorithms and Knowledge}

Human knowledge is generally provided to structure learning algorithms before they start; that is, \textit{predefined knowledge}. This predefined knowledge can, for example, be a set of directed edges that the human believes should be in the learned graph. The knowledge is often characterised as being applied as either \textit{hard} or \textit{soft constraints}. Hard constraints are those which the algorithm \textit{must} ensure the learned graph is consistent with, whereas soft constraints provide more of a guide to the learning process but which the final learned graph need not be consistent with.

One form of hard constraint is to define an ordering of the nodes in the learned graph such that the children of any node must be further down the ordering than that node. Indeed, early score-based algorithms such as K2 \citep{cooper1992bayesian} require that such a constraint is specified to reduce the search space for the algorithm. Perhaps the simplest and most widely-used form of hard constraint is to specify a set of edges, directed or otherwise, that must be in, or cannot be in the learned graph. \cite{de2007bayesian} explore this form of knowledge applied to the constraint-based PC and approximate score-based HC algorithms.

Work by \cite{chen2016learning} and \cite{wang2021learning} extends this concept by supporting ancestral constraints rather than constraints on individual arcs. \cite{borboudakis2012incorporating} focus on incorporating ancestral constraints into CPDAGs to resolve edge orientations. Recent work by \cite{brouillard2022typing} provides for variables to be grouped into different categories, for example, demographic variables, with constraints placed on the arc orientations between the various types.

Soft constraints are usually associated with score-based approaches. The prior beliefs associated with Bayesian scores provide a natural form of soft constraints \citep{heckerman1995learning}, though the huge number of possible DAGs for even a modest number of variables \citep{robinson1977counting} presents a practical challenge to defining priors on an individual DAG basis. \cite{castelo2000priors} address this issue by supporting priors on a subset of edges, and \cite{borboudakis2013scoring} allows priors to be placed on ancestral relationships between variables. \cite{amirkhani2016exploiting} implement soft constraints as an extra component within a modified BDeu score which reflects human beliefs about individual arcs, but also each human's reliability.

The authors in the above papers typically report that knowledge improves the accuracy of the learnt structure, though the simulations undertaken and metrics used vary considerably making comparisons between approaches difficult. \cite{constantinou2023impact} provide a comparison of the effectiveness of ten kinds of constraints, both soft and hard, with five algorithms. They find that predefined required arc knowledge generally improves the accuracy of the learned structure the most. With that form of knowledge, they report reductions in SHD of around 20\% when 20\% of the true arcs are predefined, rising to a 60\% reduction when half of them are predefined. They found that the approach of specifying required edges but not their orientations had the second largest impact.

The approaches discussed so far involve supplying predefined knowledge to the algorithm before it starts, but recent work includes techniques where algorithms suggest which knowledge might be most valuable. We refer to this as \textit{active learning}. One area of work relates to algorithms which attempt to identify the optimal sequence of interventional experiments that will resolve orientations in a previously learned CPDAG \citep{he2008active,li2009active}.

In contrast, \cite{murphy2001active} starts with the observational data itself and uses that to identify optimal interventional experiments using a score-based Markov Chain Monte Carlo (MCMC) sampling algorithm. \cite{statnikov2015ultra} has a similar aim but employs a constraint-based algorithm on large-scale models with up to one thousand variables. \cite{dasarathy2016active} also concentrate on large-scale problems, particularly in biology, but their algorithm requests extra observational data in areas of the graphical structure that the algorithm judges are most uncertain. ActiveBNSL \citep{ben2022active} also iteratively requests data for subsets of variables for uncertain areas of the network, making use of the GOBNILP algorithm to perform the structure learning. ActiveBNSL aims to learn an equivalence class whose score is close to that of the optimal graph for the \textit{true} distribution using data samples as efficiently as possible. Simulations with small networks of up to 12 variables demonstrated that the sample sizes required to achieve a specific accuracy could be reduced by up to 6 times with this targeted sampling. 

The active learning technique of \cite{cano2011method} is closer to the one proposed here. A score-based MCMC structure learning algorithm interactively asks a person for advice in uncertain areas of the structure - in this case, edges with a probability of existing of close to 0.5. Structural errors were reduced by around one quarter with modest numbers (10-20) of queries but the approach had the restriction that the user had to specify a predefined causal order for the variables. The algorithms in \cite{masegosa2013interactive} remove this restriction and learn a distribution of graph structures in three steps: learning the graph skeletons, then DAGs constrained to these skeletons, and finally allowing the addition of new edges to the DAGs. Human interaction is allowed at each stage with the algorithm identifying the node or edge where human input would give the greatest information gain. Simulations with networks of between 23 and 56 nodes show SHD reductions of the order of around 3 on average at a sample of 1,000 and approximately 1.5 at 5,000 rows requiring around 6 and 3 human responses respectively.

\section{Implementing Active Learning using Tabu-AL}
\label{sect:implementation}

We modify the Tabu algorithm to create the Tabu-AL algorithm which implements active learning. Tabu is widely used, competitive and relatively simple. Despite being a greedy and approximate algorithm, Tabu often provides state-of-the-art structural accuracy. It fares well in comparative studies such as that by \cite{scutari2019learns} which compared 8 score, constraint and hybrid algorithms learning from noiseless synthetic data. Tabu was most accurate in 18/20 of the case studies using discrete variable networks and the BIC score. Similarly, \cite{constantinou2021large} compared 15 algorithms including an exact score-based one, GOBNILP, and found that HC and Tabu were the most accurate learning from synthetic data both with and without different forms of noise. Since Tabu without knowledge already provides good performance, there is less room for improvement from introducing knowledge and so choosing Tabu helps to mitigate against overestimating the benefits of knowledge. We use the BIC score throughout as it is commonly used and produces good results \citep{scutari2016empirical} without the need to specify any arbitrary parameter.

\subsection{The Tabu-AL algorithm}

\algblock{Input}{EndInput}
\algnotext{EndInput}
\algblock{Output}{EndOutput}
\algnotext{EndOutput}
\newcommand{\Desc}[2]{\State \makebox[4em][l]{#1}#2}

\begin{algorithm}
    \caption{Tabu-AL algorithm} 
    \label{algo:tabu}
    \begin{algorithmic}[1]
    \Function{TABU-AL}{$data,reqd,stop$}
        \Input
        \Desc{$data$}{data set to learn graph from}
        \Desc{$reqd$}{predefined list of arcs which must be in learned graph}
        \Desc{$stop$}{predefined list of arcs that cannot be in learned graph}
        \EndInput
        \Output
        \Desc{$best\_dag$}{highest scoring DAG visited in search}
        \EndOutput
        \Statex
        \State $dag$ $\gets$ DAG containing only $reqd$ arcs  \Comment{initialise DAG}
        \State $tabulist \gets$ empty list  \Comment{fixed length list of DAGs last visited}
        \Statex
        \State $best\_dag \gets None$
        \Repeat
            \State $best\_change \gets max\_delta \gets None$
            \ForAll{allowed $dag\_change$} 
                \State $delta \gets ComputeDelta(dag\_change, dag, data)$
                \If{$delta > max\_delta$}
                    \State $max\_delta \gets delta$
                    \State $best\_change \gets dag\_change$
                \EndIf
            \EndFor
            \Statex
            \If{$IsKnowledgeRequired(best\_change, dag, data$)}
                \If {\textbf{not} $HumanSaysChangeCorrect(best\_change$)}
                    \State update $reqd$ and/or $stop$ appropriately
                    \State \textbf{break}  \Comment go to line 10 to start new iteration
                \EndIf
            \EndIf
            \Statex
            \State $dag \gets dag + best\_change$
            \State $tabulist \gets tabulist + dag$
            \If{$Score(dag, data) > Score(best\_dag, data)$}
                \State $best\_dag \gets dag$
            \EndIf
            \Statex
        \Until{stop condition}  \Comment{e.g. x iterations since last score increase}
        \Statex
        \State \textbf{return} $best\_dag$
            
        \EndFunction
	\end{algorithmic} 
\end{algorithm}

Algorithm~\ref{algo:tabu} shows the pseudo-code for the Tabu-AL algorithm. The \textit{TABU-AL} function takes the data set to learn from, \textit{data}, as an input argument, as well as \textit{reqd} and \textit{stop} arguments that allow the specification of traditional predefined required and prohibited arc constraints. \textit{dag} is the learned graph during the learning process and this is initialised as the empty graph, but with any predefined required arcs then added in.

The main loop is between lines 10 and 30 and follows the basic standard form for the HC and Tabu algorithms. In each iteration, the highest-scoring single change - an arc add, delete or reverse - is identified in lines 12 to 19. The change in score associated with each DAG change is termed the score \textit{delta}. Note that changes which would create a cycle, or a DAG in the \textit{tabulist} or violate the constraints in \textit{reqd} and \textit{stop} are not considered. This highest scoring change, \textit{best\_change}, is applied to the DAG, and the resulting DAG is added to the \textit{tabulist} in lines 26 and 27. If this DAG is also the highest-scoring DAG so far encountered, then it is also recorded as such in lines 28 to 30. The main loop terminates when a stop condition is met. In the HC algorithm, the stop condition would be when \textit{max\_delta} for that iteration is less than or equal to zero. However, the Tabu algorithm permits changes with negative deltas, and so the stop condition used is that there have been ten iterations in a row where the delta has not been positive.

The modifications which support active learning are shown in lines 20 to 25 of Algorithm~\ref{algo:tabu}. The highest scoring change to the DAG, \textit{best\_change}, having been identified, the function \textit{IsKnowledgeRequired} is called to decide if the advice of the human should be sought to see if that change is correct. This is a general approach that might encompass many kinds of criteria, but we implement and compare four criteria as described below in Subsection~\ref{sub:criteria}. If the criterion \textit{does} indicate that human advice is required, then a call to function \textit{HumanSaysChangeIsCorrect} is made to see if the human believes the proposed change is correct or not. If the human does believe the change is incorrect, then the \textit{reqd} and \textit{stop} lists are updated appropriately to prevent that change from being considered in subsequent iterations, and this iteration is terminated without the change being made to the DAG. Alternatively, if the human signals that the change is correct, then the DAG is updated as normal.

In a production system, the function \textit{HumanSaysChangeIsCorrect} might prompt a human while the algorithm is running. In that scenario, the structure learning process becomes an interactive, possibly exploratory, interaction between the algorithm and the human. Alternatively, the algorithm could pose the question to the human but then terminate, allowing some offline research or experimentation to be performed to answer the question. The algorithm could then be rerun with the researched answer included as predefined knowledge. However, for the evaluation in this paper, we simulate the human by a function described in Subsection~\ref{sub:simulatedhuman}.

\subsection{Criteria for requesting human advice}
\label{sub:criteria}

\begin{table}[H]
\scriptsize
\centering
\begin{tabular}{l  p{5cm}  p{4cm}}
\hline
\thead{Criterion name} & \thead{Criterion which highest \\ scoring change meets} & \thead{Effect of \textit{threshold} value}  \\
\hline
\textit{equivalent add} & 
It is the addition of an arc where the addition of the oppositely orientated arc is possible and has the same score. &
Not relevant for this criterion \\
\textit{small counts} & 
Associated score delta is based on contingency table(s) where a large proportion of cells have a sample count of $\leq 5$ &
Request is made if the proportion of cells with sample count $\leq 5$ is \textbf{above} the \textit{threshold} \\
\textit{unreliable score} & 
The BIC scores computed from the first and second half of the data set differ significantly, suggesting the score delta for the change might be unreliable &
Request is made when the difference between the two sub-sample scores divided by the total score is \textbf{above} the \textit{threshold} \\
\textit{small delta} & 
The score delta associated with this change is relatively small&
Request is made when score delta divided by the largest delta encountered so far is \textbf{below} the \textit{threshold} \\
\hline
\end{tabular}
\caption{Summary of the criteria used to trigger requests to the human.}
\label{tab:criteria}
\end{table}

We implement four different criteria in the \textit{IsKnowledgeRequired} function to indicate that the change proposed by the algorithm is in some senses \say{questionable} and should be referred to the human for validation. These criteria are summarised in Table~\ref{tab:criteria}. The first criterion is when the highest scoring change is to add an arc where adding the oppositely-orientated arc is possible and has the same score delta. In this case, the algorithm conventionally arbitrarily chooses one orientation or the other. This is therefore a natural circumstance in which to ask for human guidance. We refer to this criterion as an \textit{equivalent add} since the two DAGs resulting from adding either of the oppositely-orientated arcs are in the same equivalence class.

From the perspective of the graphical structure of the DAG, \textit{Equivalent add} arises where the two endpoints of the proposed arc currently have the same parents, including the case where both endpoints currently have no parents. If Tabu is starting from an empty graph then the change proposed in the first iteration will always meet this criterion. However, since the Tabu learning process often adds isolated arcs in the early part of the learning process \citep{kitson2022impact} this criterion will often be true in many of the early iterations.

The CI tests and scores which are used by most structure learning algorithms when learning from discrete data are ultimately computed from \textit{contingency tables} which hold sample counts of the number of data rows with particular combinations of variable values\footnote{More specifically, each cell in a \textit{contingency table} for a node in a discrete BN contains the sample count of data rows where that node and its parents have a specific combination of values.}. Cells with a sample count of less than 5 are generally considered an unreliable basis for statistical tests \citep{cochran1952chi2}, and therefore structure learning algorithms often disregard them \citep{spirtes2000causation,tsamardinos2006max,gasse2014hybrid}. Accordingly, we implement the \textit{small counts} criterion which detects when a large proportion of the cells underlying the change's delta have a sample count of less than 5. The relevant contingency tables are those that relate to nodes whose parents will be altered by the proposed change. For an arc reversal, these are the contingency tables relating to both endpoints of the arc, whereas only the table relating to the arc arrowhead is relevant for arc adds and deletes.

The sensitivity of this criterion is controlled by a \textit{threshold} hyperparameter which varies between 0.0 and 1.0. If the proportion of the cells which have sample counts below 5 exceeds the threshold value, then this triggers a request to the human. This hyperparameter is also used in the two further criteria described below to control their sensitivity, but note that it is applied in different ways for each criterion so that the numerical \textit{threshold} values used are not comparable across the different criteria.

Our third criterion, referred to as \textit{unreliable score}, computes the fractional difference between the BIC scores based on either the first or the second half of the data set. This criterion is based on the intuition that if the two BIC scores from subsamples of the data are rather different, this may indicate that the sample size is too small for the score deltas to be a reliable reflection of the local graph structure. If the difference between the subsample scores expressed as a proportion of the total score is above the \textit{threshold} parameter, the score is considered unreliable and the human is consulted.

The final criterion investigated, \textit{small delta}, identifies changes with relatively low deltas. These tend to be the changes near the end of the Tabu learning process where the algorithm is attempting to escape local maxima, and includes some changes which are eventually reflected in the highest-scoring graph returned, and some of which are not. Thus, this criterion identifies DAG changes which are near the decision boundary of which arcs are included in the learned graph. The BIC score is affected by the sample size and graph complexity so its absolute value will vary considerably with different networks and sample sizes. Therefore, we normalise the deltas by dividing them by the delta from the first iteration. Changes with normalised deltas below the \textit{threshold} parameter are deemed relatively small and advice from the human is requested.

\subsection{Simulating the human}
\label{sub:simulatedhuman}

The human is simulated by the function \textit{HumanSaysChangeCorrect} which uses the relevant data-generating graph to decide whether a proposed change is correct or not. Two values which control the operation of the simulated human are examined. Firstly, the \textit{limit} hyperparameter places a limit on the number of requests for knowledge that the human will answer. Once this limit is reached, the human is no longer consulted and the change proposed by the Tabu algorithm goes ahead anyway. The limit is specified as a proportion of the number of variables in the network on the basis that it is reasonable to expect that the amount of knowledge needed will rise with the number of variables and that this number is readily known in a practical setting, unlike, for instance, the number of arcs in the data generating graph. This parameter allows us to examine the effect of the amount of knowledge provided.

The second value is \textit{expertise} which defines the proportion of questions to which the simulated human gives the correct answer. The order in which correct and incorrect answers are given is randomised. So, for example, if \textit{expertise} is set to 0.8, each response has a 0.8 chance of being correct and a 0.2 chance of being incorrect. Suppose that the Tabu algorithm is proposing to add arc \(A \longrightarrow B\) which is in the data-generating graph and it is randomly chosen that the correct answer should be given. In that case, the proposed change is allowed through, and the arc is dynamically added to the \textit{reqd} list of required arcs meaning that a change which deletes or reverses that arc will not be considered in subsequent iterations. Alternatively, if it is randomly chosen that an incorrect answer be given, then the proposed change is blocked, and the \textit{reqd} and \textit{stop} constraints are updated as if either the edge was not in the graph, or the opposing arc was, again randomly decided. Note that each simulated human request is counted against the \textit{limit} regardless of whether the change is blocked or not, and whether a correct or incorrect response is given.

\section{Evaluation Methodology}
\label{sect:evaluation}

\renewcommand{\arraystretch}{1.0}  %% reduce row height of table contents
\renewcommand\theadset{\renewcommand\arraystretch{1.0}\setlength\extrarowheight{0pt}}  %% reduce row height of table headers

\begin{table}[H]
\centering
\begin{tabular}{lccccccc}
\hline
\thead{Network} & \thead{Number \\ of \\ variables} & \thead{Number \\ of \\arcs} & \thead{Mean \\ in-degree} & 
\thead{Maximum \\ in-degree} & \thead{Mean \\ degree} & \thead{Maximum \\ degree} \\
\hline
asia & 8 & 8 & 1 & 2 & 2 & 4\\
sports & 9 & 15 & 1.67 & 2 & 3.33 & 7\\
sachs & 11 & 17 & 1.55 & 3 & 3.09 & 7\\
child & 20 & 25 & 1.25 & 2 & 2.5 & 8\\
insurance & 27 & 52 & 1.93 & 3 & 3.85 & 9\\
property & 27 & 31 & 1.15 & 3 & 2.3 & 6\\
diarrhoea & 28 & 68 & 2.43 & 8 & 4.86 & 17\\
water & 32 & 66 & 2.06 & 5 & 4.12 & 8\\
mildew & 35 & 46 & 1.31 & 3 & 2.63 & 5\\
alarm & 37 & 46 & 1.24 & 4 & 2.49 & 6\\
barley & 48 & 84 & 1.75 & 4 & 3.5 & 8\\
hailfinder & 56 & 66 & 1.18 & 4 & 2.36 & 17\\
hepar2 & 70 & 123 & 1.76 & 6 & 3.51 & 19\\
win95pts & 76 & 112 & 1.47 & 7 & 2.95 & 10\\
formed & 88 & 138 & 1.57 & 6 & 3.14 & 11\\
pathfinder & 109 & 195 & 1.79 & 5 & 3.58 & 106\\
\hline
\end{tabular}
\caption{Networks used in this study}\label{tab:networks}
\end{table}

We examine the effectiveness of our active learning approach by investigating the structural accuracy of graphs learned from synthetic data generated from 16 discrete BNs commonly used in the literature. The networks are described in Table~\ref{tab:networks}, and have between 8 and 109 variables, with a variety of mean and maximum in-degrees and degrees which typify a range of different structures that one might encounter in practice. The majority of the networks are obtained from the bnlearn repository \citep{bnrepository}, but the Diarrhoea, Formed, Property and Sports networks come from the Bayesys repository \citep{bayesysrepository}.

The synthetic datasets are randomly generated according to the graphical structure and CPTs of each of the 16 networks\footnote{We compare active learning with other algorithms in subsection~\ref{sub:algos_compare} and some of these algorithms have a requirement that no variables are single-valued. We therefore slightly modify the CPT entries for a small number of variables for Water, Barley, Win95pts and Pathfinder to reduce the risk of single-valued variables}. The learning algorithms are run on subsets of this data with sample sizes of \{10\textsuperscript{3}, 5x10\textsuperscript{3}, 10\textsuperscript{4}, 5x10\textsuperscript{4}, 10\textsuperscript{5}\} which represent a range of data set sizes that might be encountered in practice.

Previous work \citep{kitson2022impact} has shown that Tabu, HC, and hybrid algorithms which make use of them, are rather sensitive to the column order of the variables in the data set. That work also found that some constraint-based algorithms were sensitive to variable order, notably the GS algorithm \citep{margaritis1999bayesian}, but also PC-Stable and IAMB \citep{tsamardinos2003time} algorithms, but to a smaller extent than HC and Tabu. To mitigate against any bias this may introduce into our results, we repeat the experiment for each network and sample size using ten different random orderings of the columns within the data set. The F1 for a particular sample size and network is obtained by taking the mean over these different orderings. Whilst this number of random orderings might be considered low, we find that the difference between our results using five or ten random orderings is small, and so ten represents a practical compromise between computing resources used and unbiased results.

We compare the effectiveness of active learning with several forms of predefined knowledge: required arcs, prohibited arcs and tiered constraints \citep{constantinou2023impact}. Analogously to active learning, the \textit{limit} value defines the number of predefined constraints used, and \textit{expertise} the proportion of those that are in agreement with the data generating graph. For example, if \textit{expertise} is set to 0.8, then 0.8 of required arcs will be randomly selected from arcs that are in the data-generating graph, but 0.2 from those that are not in the graph.

We evaluate the accuracy of the learned graphs through the widely-used F1 metric which has the advantage of being comparable across networks with different numbers of variables. As we noted in Section~\ref{sect:background}, structure learning algorithms can only determine the graphical structure up to an equivalence class represented by a CPDAG, and so structural accuracy is often assessed by comparing the learned CPDAG with the CPDAG of the data-generating (true) graph. However, since we are injecting additional edge orientation information through human knowledge, and we intend to assess the effectiveness of active learning as part of a process for learning a graph which might then be used for causal inference, our focus will be on comparing the learned and true DAGs. 

\section{Results}
\label{sect:results}

\subsection {Comparing the different criteria for requesting human advice}
\label{sub:criteria-results}

\begin{table}[H]
\centering
\begin{tabular}{lcccc}
\hline
\thead{Criteria} & \thead{Threshold \\ = 0.20} & \thead{Threshold \\ = 0.05} & 
\thead{Threshold \\ = 0.01} & \thead{Threshold \\ = 0.001} \\
\hline
small counts & 0.085 & 0.146 & 0.143 & 0.143 \\
unreliable score & 0.003 & 0.027 & 0.093 & 0.162 \\
small delta & 0.102 & 0.078 & 0.053 & 0.043 \\
\hline
equivalent add & \multicolumn{4}{c}{0.207} \\
\hline
\end{tabular}
\caption{Mean improvement in DAG F1 score over all sample sizes and networks using active learning for the four different criteria for requesting human knowledge.}
\label{tab:criteriaresults}
\end{table}

Table~\ref{tab:criteriaresults} presents the effectiveness of the four different criteria for triggering requests to the human which were described in Subsection~\ref{sub:criteria}. It shows the mean improvement in the DAG F1 score over all sixteen networks, five sample sizes, three random variable orderings, and at four different values of the \textit{threshold} value. The \textit{equivalent add} criterion does not depend upon a \textit{threshold} value and so just a single result value is shown. In all cases, the \textit{limit} hyperparameter is set to 0.50 so that no further requests are made to the human once the number of requests exceeds 0.5 times the number of variables, and all responses given by the simulated human are correct.

The results show that learned graph accuracy is improved the most by active learning when the \textit{equivalent add} criterion is used to trigger requests to the human. The mean improvement in DAG F1 accuracy is 0.207. The two criteria related to low sample size, \textit{small counts} and \textit{unreliable score}, have maximum F1 improvements of 0.146 and 0.162 respectively. The \textit{small delta} criterion was the least effective but nonetheless gave a maximum DAG F1 improvement of 0.102.

Table~\ref{tab:criteriarequests} presents an analysis of the proportion of learning iterations that result in a request being made to the simulated human for the four criteria investigated. We see that approximately a quarter of iterations result in an active learning request, with the \textit{small delta} and \textit{equivalent add} criteria making slightly fewer requests. The last two columns in this table illustrate the extent to which active learning is just providing orientation information to the algorithm. The third column shows the proportion of active learning requests that relate to edges where the Tabu algorithm has identified a true edge, and where the active learning is just confirming or correcting the orientation of that edge. We see that approximately 95\% of active learning requests fall into this category for the \textit{equivalent add} criterion. In other words, the great majority of additional information provided by active learning using the \textit{equivalent add} criterion is orientation information.

\begin{table}[H]
\centering
\begin{tabular}{lccc}
\hline
\thead{Criteria} & \thead{Rate of active \\ learning requests}
& \thead{Rate of orientation- \\ only requests}
& \thead{Rate of edge existence \\ requests} \\
\hline
small counts & 0.275 & 0.816 & 0.184 \\
unreliable score & 0.279 & 0.879 & 0.121 \\
small delta & 0.244 & 0.632 & 0.368 \\
equivalent add & 0.243 & 0.945 & 0.055 \\
\hline
\end{tabular}
\caption{Rate of active learning requests is the fraction of iterations which result in a request for knowledge being made to the human. Rate of orientation-only requests is the fraction of those active learning requests which simply correct or confirm the orientation of an edge in the true graph. Rate of existence requests is the remaining fraction of active learning requests that additionally require knowledge about the existence of the edge.}
\label{tab:criteriarequests}
\end{table}

To manage the scale of the experiments, the results presented in the following subsections only use the \textit{equivalent add} criterion since it improves accuracy the most, has a clear rationale behind it, nearly always simply supplies orientation information and does not depend on a \textit{threshold} parameter. 

\subsection{Number of active learning requests}
\label{sub:limit}

\begin{figure}[htp]
    \centering
    \includegraphics[width=13cm]{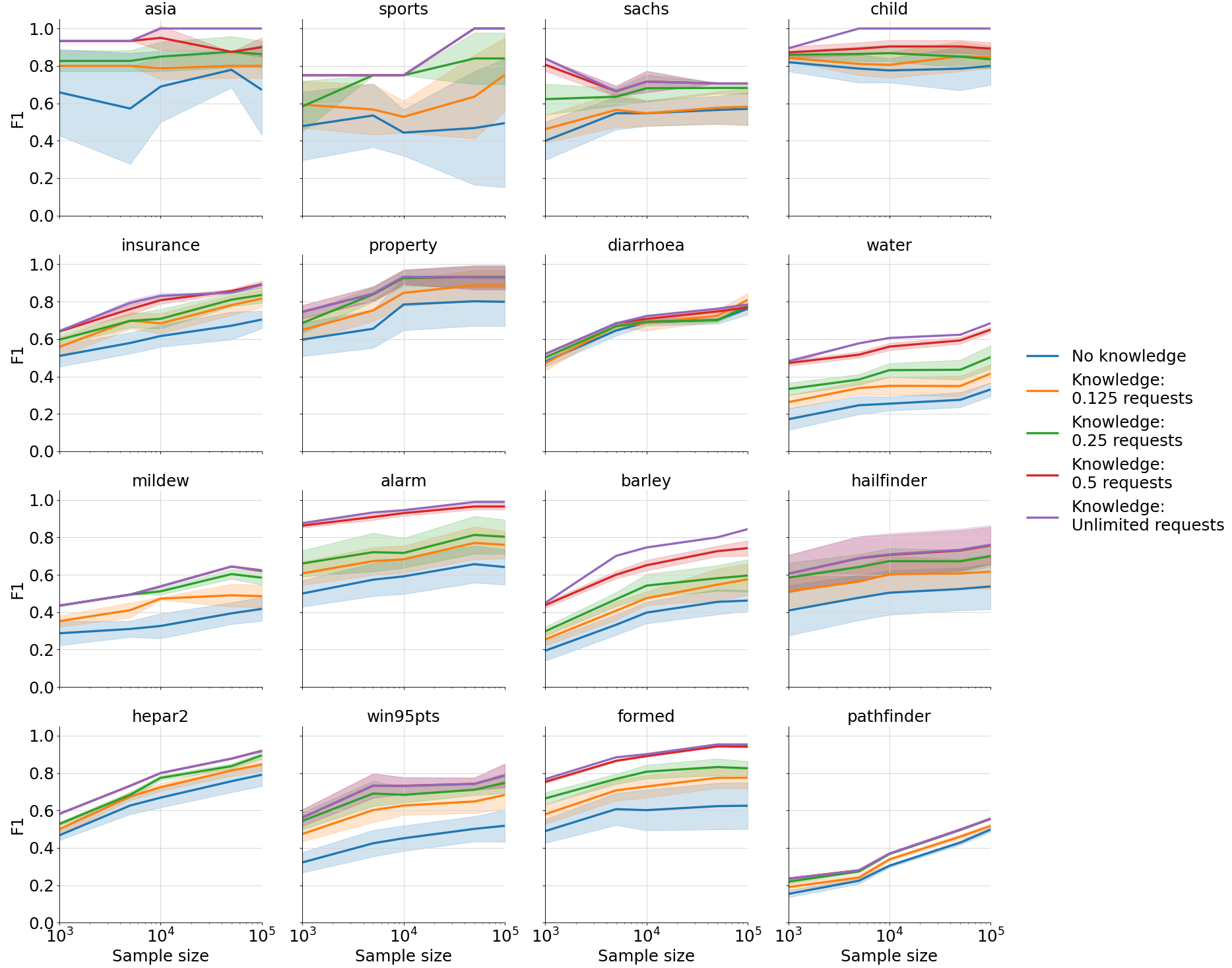}
    \caption{DAG F1 score against sample size for each network with different limits on the number of knowledge requests.}
    \label{fig:n-limit}
\end{figure}

Here we vary the \textit{limit} hyperparameter to investigate the effect that the amount of knowledge provided has on the accuracy gain. Figure~\ref{fig:n-limit} shows the DAG F1 accuracy plotted against the sample size for each of the sixteen networks investigated. For each network, we plot a baseline accuracy with no knowledge (blue line), and then the F1 with active learning knowledge with an increasing limit on the number of active learning requests allowed: 0.125, 0.25 and 0.5 times the number of variables, $n$, in the network. We also include the case where no limit is placed on the number of active learning requests. Ten experiments with different random variable orderings are performed for each combination of sample size, request limit and network. This means each line on each chart represents 50 experiments: 5 sample sizes, each with 10 variable orderings. The shaded area around each line shows the spread of F1 values over the ten different variable orderings with the upper and lower edges of the band indicating one standard deviation away from the mean value. 

As expected, we see that for nearly all sample sizes and networks, the use of active learning improves the mean accuracy over the ten variable orderings. Increasing the amount of knowledge generally increases the accuracy gain. Similarly, we see an expected upward trend in accuracy as the sample size grows. For most networks, except Barley and Child, the red and purple lines are close, indicating that nearly all the possible benefit from active learning is obtained when the number of active learning requests is equal to $0.5 \times n$.

Table~\ref{tab:limit} summarises the mean change in F1 resulting from active learning for each combination of network and limit on the number of active learning requests. For each experiment with a particular variable ordering, sample size and network, the improvement in F1 is calculated by subtracting the F1 achieved with no active learning from the same experiment when active learning is used. These individual F1 score differences are then averaged across all sample sizes and variable orderings to give a mean F1 change for each combination of network and limit on the number of active learning requests shown in Table~\ref{tab:limit}. Table~\ref{tab:limit-shd} in \ref{app:shd} presents the same results using the SHD metric.

The mean change in F1 is positive for every network and amount of active learning requests, and the improvement generally increases as the limit on the number of requests is raised. The improvement in F1 is often considerable even when the limit is set at $0.125 \times n$. For example, F1 improves by 0.15 with just 2 requests for the Sports network, and 0.163 with 19 requests on the larger Win95pts network. As noted already, a limit of $0.5 \times n$ usually achieves a similar accuracy improvement to no limit and so represents a useful \say{rule of thumb} for the amount of knowledge required to achieve close to maximum benefit. The bottom line of Table~\ref{tab:limit} shows the mean F1 improvement over all networks which is considerable at the limits tested. Nonetheless, there are some networks, in particular Diarrhoea and Pathfinder, where active knowledge has a much smaller benefit.

\begin{table}[H]
\centering
\begin{tabular}{lccccc}
\hline
\thead{Network} & \thead{Number of \\ variables, $n$} & \thead{Limit \\ $0.125 \times n$} & \thead{Limit \\ $0.25 \times n$} & \thead{Limit \\ $0.5 \times n$} & \thead{No limit} \\
\hline
asia & 8 & 0.123 & 0.174 & 0.244 & 0.299 \\
sports & 9 & 0.132 & 0.269 & 0.366 & 0.366 \\
sachs & 11 & 0.021 & 0.135 & 0.194 & 0.201 \\
child & 20 & 0.038 & 0.062 & 0.100 & 0.186 \\
insurance & 27 & 0.092 & 0.113 & 0.175 & 0.185 \\
property & 27 & 0.077 & 0.135 & 0.148 & 0.148 \\
diarrhoea & 28 & 0.016 & 0.012 & 0.030 & 0.039 \\
water & 32 & 0.087 & 0.162 & 0.303 & 0.339 \\
mildew & 35 & 0.095 & 0.179 & 0.199 & 0.200 \\
alarm & 37 & 0.106 & 0.150 & 0.334 & 0.354 \\
barley & 48 & 0.084 & 0.129 & 0.263 & 0.340 \\
hailfinder & 56 & 0.090 & 0.164 & 0.207 & 0.209 \\
hepar2 & 70 & 0.050 & 0.082 & 0.120 & 0.120 \\
win95pts & 76 & 0.163 & 0.232 & 0.267 & 0.267 \\
formed & 88 & 0.124 & 0.190 & 0.289 & 0.302 \\
pathfinder & 109 & 0.028 & 0.060 & 0.066 & 0.066 \\
\hline
\multicolumn{2}{c}{ALL NETWORKS} & 0.083 & 0.140 & 0.207 & 0.226 \\
\hline
\end{tabular}
\caption{Mean improvement in DAG F1 score over all the sample sizes using active learning for each network and differing limits on the number of requests.}
\label{tab:limit}
\end{table}

The results in Table~\ref{tab:ordering_sd} confirm that increasing the amount of knowledge has a beneficial side-effect of reducing the Tabu algorithm's sensitivity to variable ordering. The standard deviation of the F1 score over the ten variable orderings for each combination of sample size, network and the limit on active learning requests is first computed. Table~\ref{tab:ordering_sd} presents the mean of these standard deviations taken over all networks and sample sizes at each request limit. There is a steady trend of the mean standard deviation in F1 falling as the amount of active learning requests is increased.

\begin{table}[H]
\centering
\begin{tabular}{lc}
\hline
\thead{Limit on number of \\ active learning requests} 
& \thead{Mean Standard Deviation \\ in F1} \\
\hline
No active learning & 0.088 \\
$0.125 \times n$ & 0.059 \\
$0.25 \times n$ & 0.045 \\
$0.5 \times n$ & 0.024 \\
No limit on active learning requests & 0.015 \\
\hline
\end{tabular}
\caption{Sensitivity to variable ordering for differing limits on the number of active learning requests. The limit on the number of active learning requests is expressed as a proportion of the number of variables, $n$, in each network. The mean standard deviation in F1 is computed across the ten variable orderings and then averaged over all sample sizes and networks.}
\label{tab:ordering_sd}
\end{table}

\subsection{Comparison with predefined knowledge}
\label{sub:predefined-results}

\begin{figure}[htp]
    \centering
    \includegraphics[width=12cm]{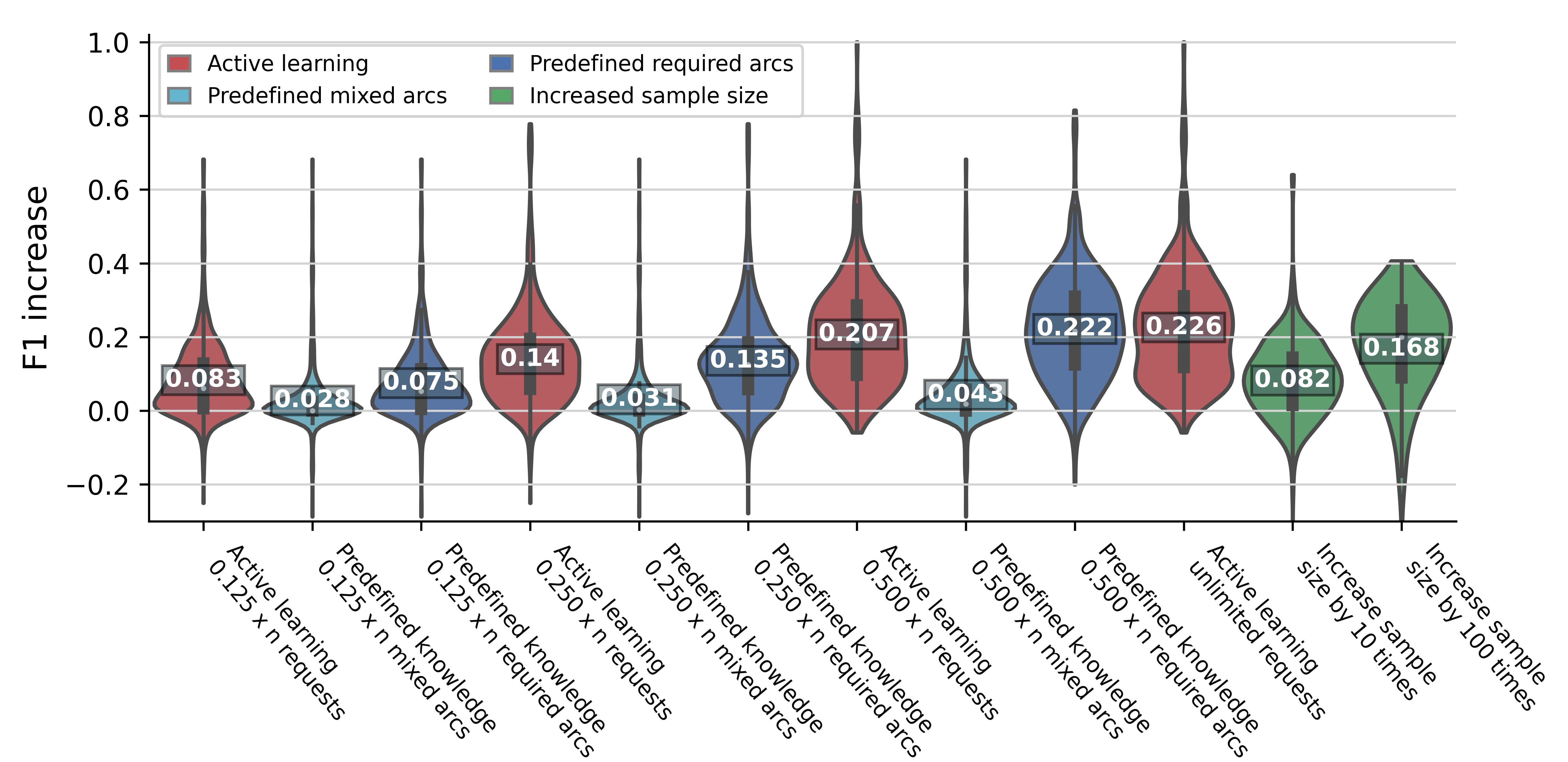}
    \caption{Distributions of DAG F1 change over no knowledge for both active learning and predefined knowledge for different limits on the number of knowledge items.}
    \label{fig:act-predef}
\end{figure}

We compare the accuracy improvements gained with active learning with more traditional predefined knowledge in Figure~\ref{fig:act-predef}. The red violin plots show the distribution in F1 improvement resulting from active learning and the blue violin plots show the improvement from predefined knowledge. As with the results in Table~\ref{tab:limit}, the improvement in F1 for each individual experiment is just the F1 score obtained with knowledge minus the F1 value when no knowledge is used.

We consider two types of predefined knowledge in this figure. The first type of predefined knowledge investigated in the figure is where only required arcs are specified. The second type is referred to as \textit{mixed arcs} and is where both prohibited and required arcs are predefined, in a ratio of 9 to 1 respectively. This is because, as discussed in Subsection~\ref{sub:criteria-results}, the great majority of active learning responses merely correct or confirm the orientation information of edges that Tabu is already adding. This \textit{mixed arcs} predefined knowledge has a similarly weighted bias towards providing mostly orientation information.

The first red \say{violin} plot in Figure~\ref{fig:act-predef} shows the accuracy gain with active learning limited to $0.125 \times n$ requests. The blue plot immediately to its right shows accuracy gains with a mixture of required and prohibited arcs predefined, with a total of $0.125 \times n$ arcs specified. The following darker blue plot shows the accuracy gain with $0.125 \times n$ predefined required arcs. Each individual plot on the figure represents the distribution of the F1 gain over 800 experiments consisting of 5 sample sizes, 10 variable orderings and 16 networks. The top and bottom of each violin plot show the maximum and minimum F1 gain over those experiments and the rectangle inside the violin indicates the interquartile range. Each violin is annotated with the mean F1 gain. This pattern of comparing the three forms of knowledge is repeated at $0.25$ and $0.5 \times n$, so that Figure~\ref{fig:act-predef} also illustrates the effect of the \textit{amount} as well as \textit{type} of knowledge supplied. The rightmost red plot in Figure~\ref{fig:act-predef} shows the accuracy gain with no limit placed on the number of active learning requests.

In order to place the improvements in F1 due to knowledge in context, Figure~\ref{fig:act-predef} also includes improvements in F1 due to increasing the sample size, shown by the green-coloured plots. These are based on the experiments when no knowledge is used and involve comparing the F1 achieved at a given sample size with the F1 achieved with a sample size ten times and one hundred times larger. This comparison is done for each network and variable ordering available. So, for example, the F1 achieved with a specific variable ordering for a particular network with a sample size of 1,000 and 10,000 is compared, and 5,000 with 50,000 and so on.

Each distribution has long tails, so there is a small proportion of combinations of network, sample size and variable ordering where knowledge generates unusually large improvements in F1. Conversely, there is a small proportion where knowledge \textit{worsens} accuracy greatly. However, we see that active learning with a limit of $0.5 \times n$ requests and unlimited requests does not produce large accuracy decreases in contrast to the other experiments.

Active learning and predefined required arcs achieve similar levels of F1 improvement at each knowledge limit, with active learning being slightly more effective at limits of $0.125 \times n$ and $0.25 \times n$, whereas predefined required arcs give better accuracy at $0.5 \times n$. We note that predefined required arcs supply the algorithm with information about the existence \textit{and} orientation of each arc specified, whereas as shown in Table~\ref{tab:criteriarequests}, active learning tends to supply only additional orientation information. Arguably, therefore, the comparison between active learning and the mixed predefined arcs which have a similar balance between orientation and existence information is fairer. Here, the accuracy gains from active learning are between 3 and 5 times that from predefined knowledge.

Increasing the sample size generally improves F1, but there is a small proportion where increasing the sample size worsens F1. The mean improvement in F1 due to increasing the sample size by ten times is 0.082, and by one hundred times is 0.168. Active learning with a limit of $0.125 \times n$ requests improves accuracy by a very similar amount to increasing the sample size by 10 times.

Given that the active learning we have considered so far mostly supplies orientation information, and that humans may be less confident in saying whether edges exist or not, we present some comparisons where only orientation knowledge is used. To achieve this, we restrict active learning so that the human is never asked to confirm whether an arc deletion should proceed; i.e., in this experiment, when an arc is being added or reversed, the human can only say whether the proposed orientation is correct or not, but cannot indicate that the arc does not exist.

We compare this orientation-only active learning with two forms of predefined knowledge that also only supply orientation information. The first comparison uses $0.125, 0.25$ or $0.5 \times n$ arcs which are \textit{not in} the true graph, and are predefined as prohibited arcs. Each prohibited arc ensures that the relevant edge can only be added in one orientation. The second approach is to assign a subset of nodes to tiers, such that arcs are prohibited from a lower tier to a higher tier, and are also prohibited within a tier. Tier-based predefined knowledge is often used to enforce temporal constraints. A random selection of $0.125, 0.25$ and $0.50 \times n$ nodes are assigned to tiers according to the topological ordering of the true graph. The set of prohibited arcs corresponding to this partial tier assignment is generated and forms the predefined knowledge used in this comparison. Note that a given number of nodes assigned to tiers will generally result in a larger number of individual prohibited arcs, and so this represents a stronger form of orientation-only predefined knowledge than when the same limit is applied to simple prohibited arcs.

\begin{figure}[htp]
    \centering
    \includegraphics[width=12cm]{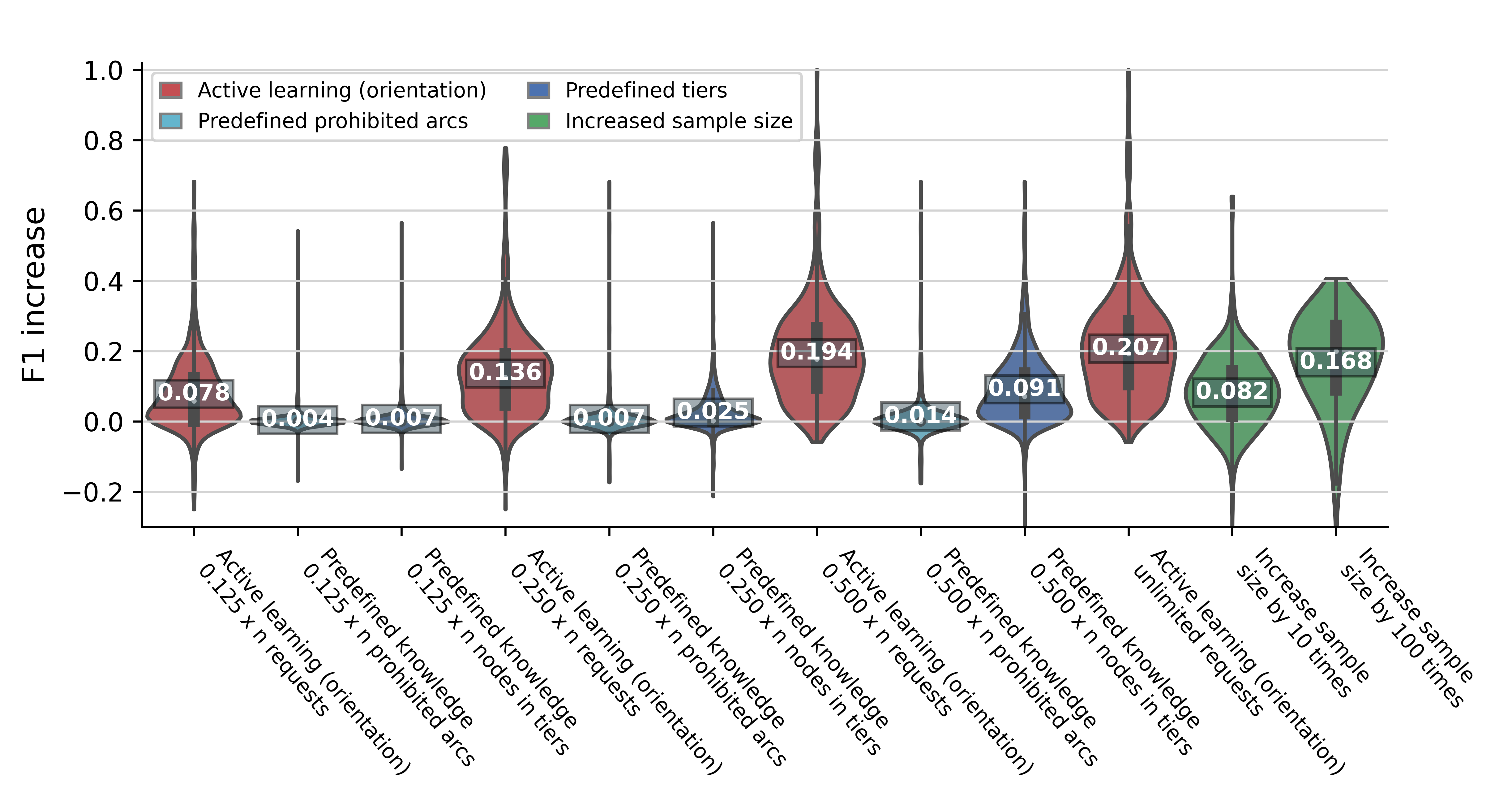}
    \caption{Distributions of DAG F1 change over no knowledge for both active learning and predefined knowledge for different limits on the number of orientation-only knowledge items.}
    \label{fig:act-predef-orient}
\end{figure}

Figure~\ref{fig:act-predef-orient} compares orientation-only active learning and predefined knowledge. Comparing the accuracy gains from active learning in Figure~\ref{fig:act-predef} with the orientation-only active learning in this figure we first observe that the improvement in F1 from orientation-only active learning is only slightly smaller than when active learning allows the human to adjudicate on edge existence. Across the request limits tested, orientation-only active learning is on average, around six per cent lower than when existence information is additionally allowed. This presumably reflects the fact that the majority of active learning requests concern orientation decisions. At a limit of $0.125 \times n$ requests, orientation-only active learning improves F1 by 0.078 compared to 0.082 resulting from increasing the sample size by ten times. With a limit of $0.5 \times n$, orientation-only active learning improves F1 by 0.194, higher than the 0.168 resulting from increasing the sample size by one hundred times. Thus, orientation-only active learning also provides an effective means to improve the accuracy of structure learning.

Figure~\ref{fig:act-predef-orient} shows that orientation-only active learning is considerably more effective than the two predefined knowledge approaches which also just supply orientation information. The accuracy gain with prohibited arcs is very modest at all knowledge amounts, increasing F1 by between approximately fourteen and twenty times less than active learning. This echoes results by \cite{constantinou2023impact} indicating that prohibited arcs have relatively little effect on learnt accuracy. Assigning a given number of nodes to tiers is more effective, but improves F1 by between around two and eleven times less than active learning. These results suggest that active learning may be particularly useful where humans only wish to specify arc orientation.

\subsection{Imperfect Knowledge}
\label{sub:imperfect}

\begin{figure}[htp]
    \centering
    \includegraphics[width=13cm]{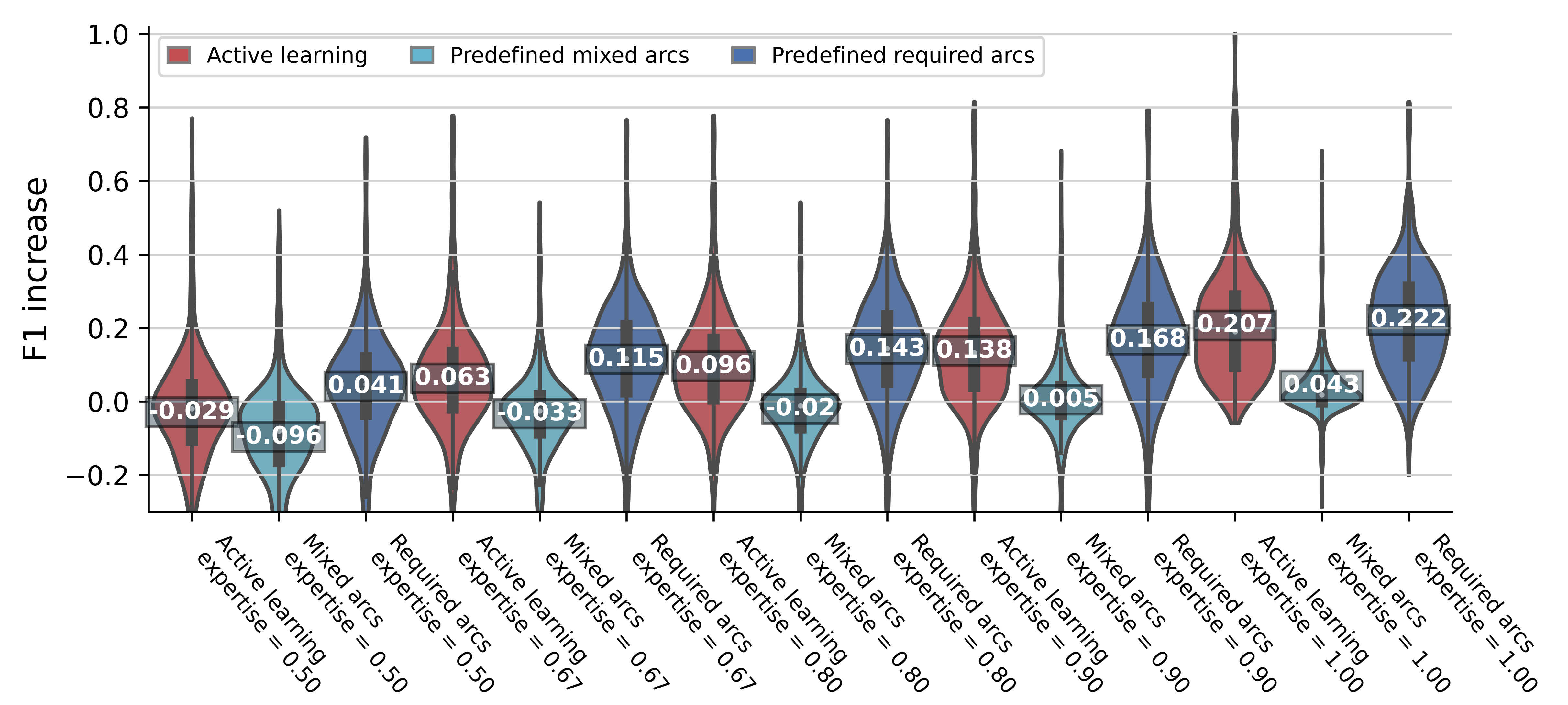}
    \caption{Distributions of DAG F1 change over no knowledge for both active learning and predefined knowledge at differing levels of expertise.}
    \label{fig:act-exp}
\end{figure}

The above results have all assumed that each piece of knowledge provided by the human is correct. In this section, we simulate a human responding incorrectly by varying the \textit{expertise} value which defines the proportion of knowledge that is correct. For example, if $expertise = 0.80$, there will be a 0.2 probability that an active learning response will be incorrect, and a 0.2 probability that a predefined required arc will not actually be in the data-generating graph. These experiments use a knowledge limit of $0.5 \times n$ and expertise values of 0.50, 0.67, 0.80, 0.90 and 1.00. Figure~\ref{fig:act-exp} shows F1 improvements at the different expertise levels for the predefined and active learning approaches presented in Figure~\ref{fig:act-predef}, The distributions for active learning are in red, and the corresponding one for the same expertise level for predefined knowledge in blue. Table~\ref{tab:exp-shd} presents the results of this experiment using active learning according to the SHD metric.

As one might expect, accuracy worsens as the expertise level is reduced. However, this degradation is very pronounced with a mix of predefined prohibited and required arcs, with a mean F1 improvement of only 0.005 at $expertise = 0.90$, and with mean F1 worsened with imperfect knowledge at lower expertise levels. Active learning fares better, though with a substantial drop from a 0.207 improvement in F1 with perfect knowledge to a 0.138 improvement at $expertise = 0.90$, and 0.096 and 0.063 at $expertise = 0.80$ and 0.67 respectively. The predefined required arcs approach is the most robust to human error, with F1 gain falling to 0.168 at $expertise = 0.90$, 0.143 at 0.80 and 0.115 at 0.67. Somewhat surprisingly, predefined required arcs even improves F1 (by 0.041) at $expertise = 0.50$ suggesting that correct required arcs might have a stronger beneficial effect than the adverse effect of incorrect required arcs.

\begin{figure}[htp]
    \centering
    \includegraphics[width=13cm]{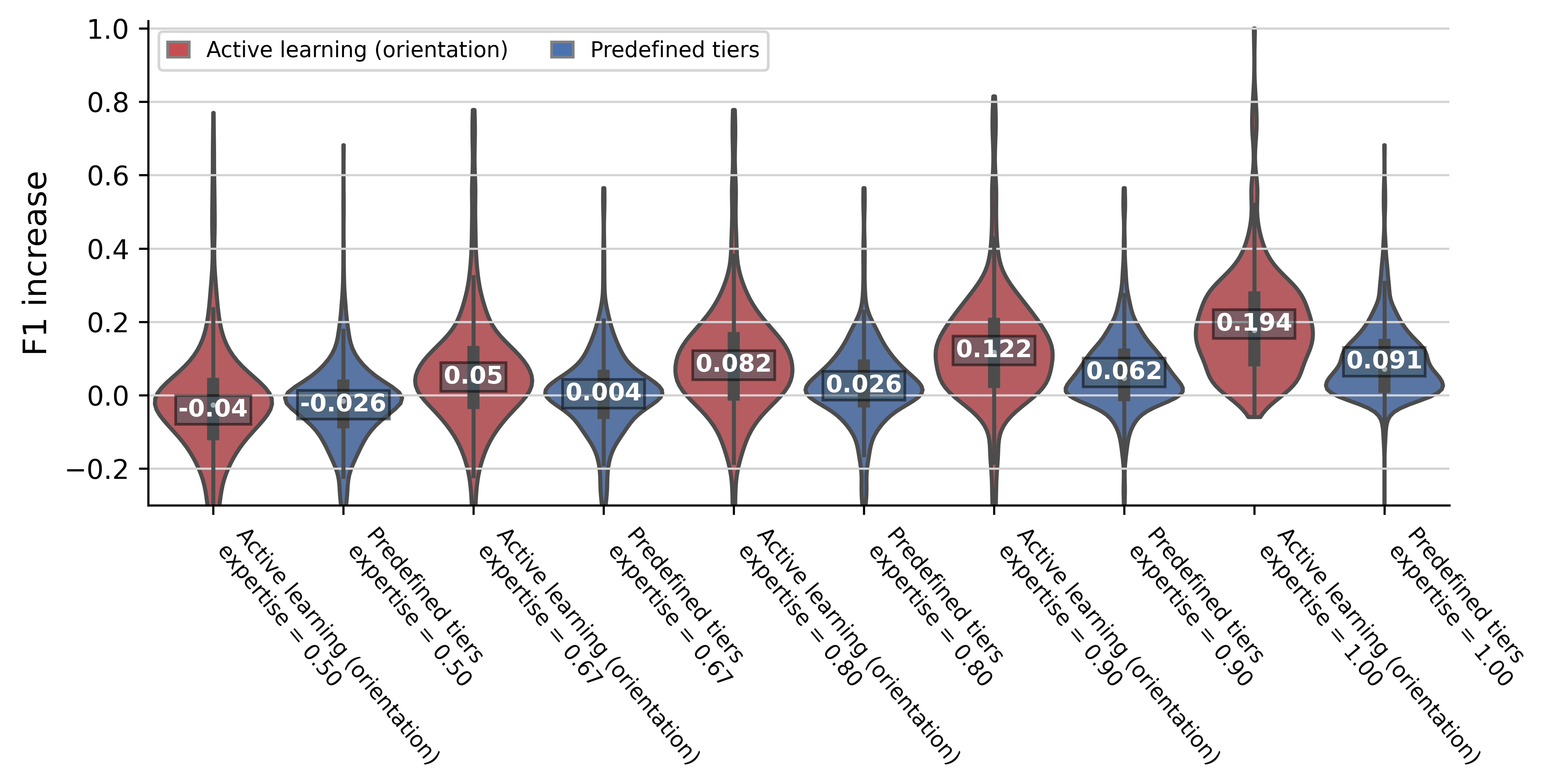}
    \caption{Distributions of DAG F1 change over no knowledge for orientation-only active learning and predefined knowledge at differing levels of expertise.}
    \label{fig:act-exp-orient}
\end{figure}

Figure~\ref{fig:act-exp-orient} compares the effect of expertise level on orientation-only predefined knowledge and active learning. Given the small beneficial effects of predefined prohibited arcs with perfect knowledge, we do not show the effect of expertise on predefined prohibited arcs here. Figure~\ref{fig:act-exp-orient} shows that F1 improvement using predefined knowledge degrades more rapidly than active learning with, for example, active learning improving F1 by ten times more than predefined tiers at $expertise = 0.67$.

\subsection{Comparisons with other algorithms}
\label{sub:algos_compare}

\begin{figure}[htp]
    \centering
    \includegraphics[width=13cm]{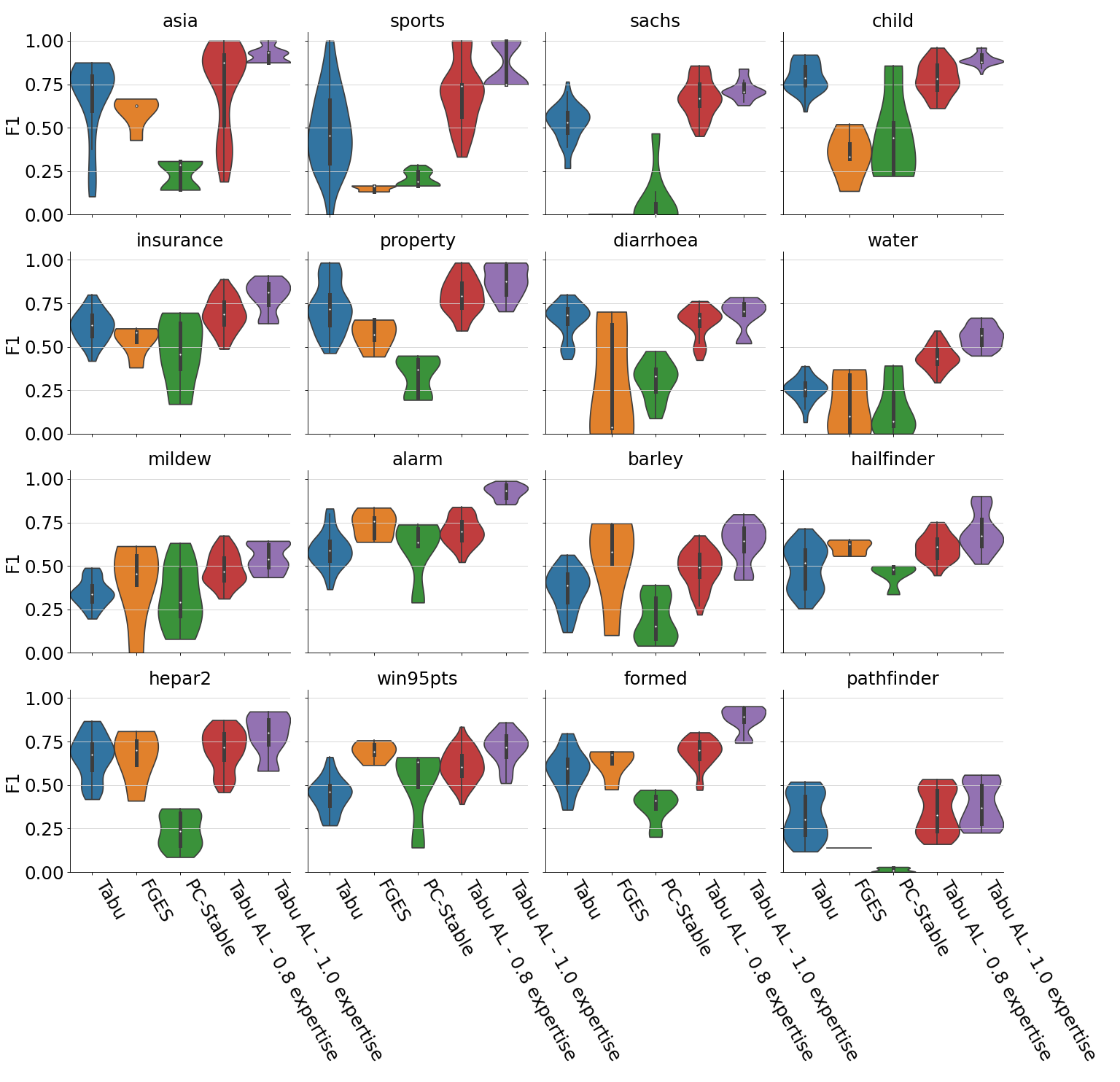}
    \caption{DAG F1 distribution for each network for Tabu with and without active learning, and the FGES and PC-Stable algorithms.}
    \label{fig:algo}
\end{figure}

This final subsection compares the DAG F1 accuracy achieved using Tabu-AL with standard Tabu and two other commonly used algorithms without knowledge, to illustrate the range of accuracies one might expect to see in structure learning in a practical setting. Figure~\ref{fig:algo} shows the distribution in F1 DAG accuracy over all the sample sizes for each network and algorithm. Time constraints meant that only three different variable orderings were used with the PC algorithm, and only one variable ordering with the FGES algorithm since previous work \citep{kitson2022impact} had shown that the latter was rather insensitive to variable ordering. Results from the orientation-only variant of active learning are shown since this may be easier to use as it demands less information from the human.

These results are not intended as a comparison of algorithm performance since we compare algorithms where knowledge is not provided with results where Tabu is aided by human knowledge. Moreover, FGES and PC-Stable produce CPDAGs rather than DAGs, and the comparison semantics we use to compute F1 penalise the case where the learned graph has an undirected edge. Nonetheless, we see that using Tabu-AL restricted to edge orientation only generally provides higher accuracy when the goal is to learn a DAG. This remains true when the human knowledge is imperfect and so we suggest that active learning is a fruitful approach in practical problems where the aim is to learn a causal graph. We also note that networks where active learning offers the smallest benefits such as Child, Diarrhoea and Pathfinder are also networks where the PC-Stable and FGES algorithms perform relatively poorly too.

\section{Conclusions and Future Work}
\label{sect:conclusions}

This paper presents the Tabu-AL structure learning algorithm where knowledge is requested from a human as the learning process proceeds - an approach known as \textit{active learning}. This is in contrast to the usual technique for combining knowledge, where knowledge is provided upfront to the algorithm without any guidance as to what knowledge might be useful or not, which we term \textit{predefined} knowledge. We use a novel approach based on the widely-used Tabu structure learning algorithm, where we add a criterion to judge whether each proposed change to the DAG may be incorrect and if so, to ask the human for confirmation. We explore four different criteria for deciding whether a change is likely to be incorrect and find the most effective criterion to be when the Tabu-AL algorithm is adding an arc at a point when it is also possible to add the opposite arc with the same objective score improvement.

We evaluate the technique by generating synthetic data from sixteen well-known BNs and assess the structural accuracy of the graph learnt from the synthetic data. Our interest is in learning the causal graph and so our primary comparison metric is the F1 metric for the learned DAG. We compare results with and without active learning, compare active learning to predefined knowledge, and investigate the sensitivity to incorrect knowledge. We simulate the human with a function that supplies answers based on the data-generating graph with a defined level of accuracy.

Active learning improves the accuracy of the learned graph considerably. Limiting the number of requests to 0.125 times the number of variables, $n$, in the model, we find a mean F1 improvement of 0.083 over using no knowledge, which rises to 0.207 when the number of requests is capped at $0.5 \times n$. These improvements compare favourably with those arising from simply increasing sample size, where increasing the sample size by ten-fold and one hundred-fold improves F1 by 0.082 and 0.168 respectively. Active learning also reduces the sensitivity of Tabu to variable ordering with the standard deviation in F1 arising from different variable orders falling from a mean of 0.088 to 0.024 with active learning capped at $0.5 \times n$.

Analysis shows that the great majority of active learning requests, typically around 95\%, relate to an edge that Tabu is correctly adding to the graph but where the orientation needs confirmation or correction. We therefore also investigate a variant of active learning where the human is only asked about the orientation of arcs in the graph being learnt. This orientation-only active learning is nearly as effective as active learning where the existence of arcs is also checked. We suggest that this form of active learning may be particularly attractive to practitioners since they may be more confident about adjudicating solely on arc orientations without the need to contradict edge existence made by the algorithm. However, we stress that practitioners must continue to be aware of the issues listed in Subsection~\ref{sub:algos} relating to causal discovery even when active learning is used, especially with noisy data.

The accuracy improvement from active learning is compared to traditional approaches using predefined knowledge. Whilst it is not possible to make an exact comparison, we find that active learning is more effective at improving accuracy than predefined knowledge where a similar amount of primarily orientation information is provided by the human. In this case, active learning improves accuracy by between three and five times more than a comparable mix of predefined required and prohibited arcs. Only predefined knowledge using required arcs has a similar effectiveness to active learning, but we note that this form of predefined knowledge represents information about both the orientation and existence of arcs, and so represents a larger amount of information supplied to the structure learning algorithm, and which may not be available in many practical applications. When orientation-only knowledge is used, active learning is much more effective than prohibited arcs or tier constraints, increasing accuracy by at least fourteen and twice times respectively.

The F1 improvement due to active learning falls as the proportion of incorrect human knowledge is increased. It drops by around one-third when 10\% of the human's responses are incorrect, and by just over a half when 20\% of responses are incorrect. Nonetheless, active learning still provides a considerable F1 gain when one-third of the responses are incorrect. Also, the effectiveness of predefined knowledge generally falls even more rapidly with increasing proportions of incorrect knowledge. The exception to this is predefined required arcs which are more robust to human error than active learning. The decreased sensitivity of required arcs to incorrect knowledge may, however, be offset by the fact that predefined required arcs are probably much harder to specify and therefore more likely to be incorrect. When restricted to orientation-only knowledge, active learning is more robust to human mistakes than predefined knowledge.

As well as improving structural accuracy, active learning also offers increased transparency and human engagement in the structure-learning process, which may in turn lead to a more efficient use of human expertise. Rather than having to think about all the variables involved upfront, active learning guides the practitioner towards knowledge that the algorithm would benefit from. Tabu-AL could be used purely interactively, or run repeatedly allowing the practitioner to research answers to questions it had posed in between runs.

This work could be usefully extended by applying active learning to other score-based or constraint-based algorithms or using it with noisy data of all forms - where there is missing data or variables, or measurement error. We also note that the questions that arise in active learning may be suitable for a Large Language Model to answer \citep{long2023can} which would be an interesting avenue to pursue. However, perhaps the most interesting exercise would be to compare this approach with other algorithms and other forms of knowledge in a real-world practical problem, ideally in a situation with domain experts available and some means of validating the learned graphs against the underlying causal model. 

\label{}



\bibliographystyle{elsarticle-harv} 
\bibliography{references}

\appendix

\section{Selected results using the SHD metric}
\label{app:shd}

We repeat some results from the main paper here, but expressed using the SHD metric which is commonly reported in other studies. This may be useful in comparing results obtained here with other studies. It bears repeating that since we are focused on learning a causal graph the comparisons here are between the learned graph and data-generating DAG.

\begin{table}[H]
\centering
\begin{tabular}{lccccc}
\hline
\thead{Network} & \thead{Number of \\ variables, $n$} & \thead{Limit \\ $0.125 \times n$} & \thead{Limit \\ $0.25 \times n$} & \thead{Limit \\ $0.5 \times n$} & \thead{No limit} \\
\hline
asia & 8 & 1.6 & 2.0 & 2.5 & 2.9 \\
sports & 9 & 2.8 & 5.3 & 6.7 & 6.7 \\
sachs & 11 & 0.4 & 2.3 & 3.3 & 3.4 \\
child & 20 & 1.2 & 2.1 & 3.2 & 5.3 \\
insurance & 27 & 6.5 & 7.7 & 11.5 & 12.0 \\
property & 27 & 2.9 & 5.0 & 5.1 & 5.1 \\
diarrhoea & 28 & 1.2 & 0.8 & 2.1 & 2.7 \\
water & 32 & 5.6 & 9.8 & 16.8 & 18.8 \\
mildew & 35 & 5.5 & 10.1 & 11.4 & 11.5 \\
alarm & 37 & 7.0 & 10.0 & 21.6 & 22.5 \\
barley & 48 & 8.1 & 12.4 & 26.6 & 34.3 \\
hailfinder & 56 & 8.2 & 15.5 & 19.8 & 20.1 \\
hepar2 & 70 & 6.7 & 11.0 & 16.0 & 16.0 \\
win95pts & 76 & 34.1 & 46.8 & 50.8 & 50.8 \\
formed & 88 & 27.8 & 39.4 & 56.5 & 58.4 \\
pathfinder & 109 & 7.1 & 15.0 & 16.5 & 16.5 \\
\hline
\end{tabular}
\caption{Mean improvement (that is, decrease) in SHD over all the sample sizes using active learning compared to no knowledge for each network and differing limits on the number of requests.}
\label{tab:limit-shd}
\end{table}

Table~\ref{tab:limit-shd} expresses the results in Table~\ref{tab:limit} using SHD rather than F1. The SHD results largely follow the trends apparent from F1.  For example, Diarrhoea and Pathfinder demonstrate a small benefit, whereas Barley and Formed see a large benefit from active learning. There are, however, some differences between the two metrics. The benefits of active learning applied to the Child network are more pronounced according to the F1 metric. Also, increasing the limit on the amount of knowledge always increased F1 but it reduces the SHD improvement in the case of Property.

\begin{table}[H]
\centering
\begin{tabular}{lcccccc}
\hline
\thead{Network} & \thead{Number of \\ variables, $n$} & \thead{0.50} & \thead{0.67} & \thead{0.80} & \thead{0.90} & \thead{1.00} \\
\hline
asia & 8 & \textcolor{red}{-1.8} & 0.1 & 0.3 & 1.2 & 2.5 \\
sports & 9 & 1.0 & 3.5 & 4.0 & 4.8 & 6.7 \\
sachs & 11 & 0.3 & 1.9 & 2.1 & 2.0 & 3.3 \\
child & 20 & \textcolor{red}{-6.0} & \textcolor{red}{-1.5} & \textcolor{red}{-1.1} & 1.2 & 3.2 \\
insurance & 27 & \textcolor{red}{-10.5} & \textcolor{red}{-0.6} & 3.4 & 6.7 & 11.5 \\
property & 27 & \textcolor{red}{-5.4} & \textcolor{red}{-0.4} & 1.5 & 2.9 & 5.1 \\
diarrhoea & 28 & \textcolor{red}{-5.9} & \textcolor{red}{-3.4} & \textcolor{red}{-2.3} & 0.2 & 2.1 \\
water & 32 & \textcolor{red}{-1.5} & 4.5 & 8.4 & 11.9 & 16.8 \\
mildew & 35 & 0.3 & 4.9 & 7.5 & 8.6 & 11.4 \\
alarm & 37 & \textcolor{red}{-10.0} & 0.4 & 4.6 & 10.4 & 21.6 \\
barley & 48 & \textcolor{red}{-3.3} & 4.2 & 9.8 & 18.1 & 26.6 \\
hailfinder & 56 & \textcolor{red}{-6.8} & 4.8 & 9.5 & 15.4 & 19.8 \\
hepar2 & 70 & \textcolor{red}{-12.7} & \textcolor{red}{-1.0} & 3.1 & 7.9 & 16.0 \\
win95pts & 76 & \textcolor{red}{-1.9} & 18.6 & 28.7 & 38.0 & 50.8 \\
formed & 88 & \textcolor{red}{-28.8} & 0.6 & 12.1 & 32.6 & 56.5 \\
pathfinder & 109 & \textcolor{red}{-0.2} & 2.0 & 5.9 & 10.6 & 16.5 \\
\hline
\multicolumn{2}{c}{networks where SHD improved:} & 3/16 & 11/16 & 14/16 & 16/16 & 16/16 \\
\hline
\end{tabular}
\caption{Mean improvement in SHD over all the sample sizes using active learning compared to no knowledge for each network and differing levels of expertise. Negative SHD improvements, which is where active learning \textit{increased} SHD, are marked in red}
\label{tab:exp-shd}
\end{table}

Table~\ref{tab:exp-shd} shows the improvement in SHD due to active learning at the levels of expertise investigated in Subsection~\ref{sub:imperfect}. The results are broken down by network here as SHD is not readily comparable across different network sizes and all relate to a request limit of $0.5 \times n$. The results are in line with the F1 ones, in that the accuracy of nearly all networks worsened with active learning at $expertise=0.5$, but the majority improved at 0.67 and 0.80, and all did at higher levels of expertise.

\end{document}